\documentclass[a4paper,fleqn]{cas-sc}

\usepackage[authoryear]{natbib}

\def\tsc#1{\csdef{#1}{\textsc{\lowercase{#1}}\xspace}}
\tsc{WGM}
\tsc{QE}
\tsc{EP}
\tsc{PMS}
\tsc{BEC}
\tsc{DE}

\begin{document}
\let\WriteBookmarks\relax
\def\floatpagepagefraction{1}
\def\textpagefraction{.001}
\shorttitle{}
\shortauthors{Z. Fang et al.}

\title [mode = title]{LPUWF-LDM: Enhanced Latent Diffusion Model for Precise Late-phase UWF-FA Generation on Limited Dataset}                      

\affiliation[1]{organization={Hangzhou Dianzi University},
                city={Hangzhou},
                postcode={310018}, 
                country={China}}

\author[1]{Zhaojie Fang}\ead{21321206@hdu.edu.cn}
\fnmark[1]
\author[1]{Xiao Yu}\ead{22320313@hdu.edu.cn}
\fnmark[1]
\author[1]{Guanyu Zhou}\ead{23320307@hdu.edu.cn}
\author[1]{Ke Zhuang}\ead{22061725@hdu.edu.cn}
\author[1]{Yifei Chen}\ead{chenyifei@hdu.edu.cn}
\author[1]{Ruiquan Ge}\ead{gespring@hdu.edu.cn}
\cormark[1]

\author[2]{Changmiao Wang}\ead{cmwangalbert@gmail.com}
\cormark[1]

\author[1]{Gangyong Jia}\ead{gangyong@hdu.edu.cn}
\author[1]{Qing Wu}\ead{wuq@hdu.edu.cn}
\author[4]{Juan Ye}\ead{yejuan@zju.edu.cn}
\author[4]{Maimaiti Nuliqiman}\ead{22218653@zju.edu.cn}
\author[4]{Peifang Xu}\ead{xpf1900@zju.edu.cn}
\author[3]{Ahmed\ Elazab}\ead{Ahmedelazab@szu.edu.cn}

\affiliation[2]{organization={Shenzhen Research Institute of Big Data},
                city={Shenzhen},
                postcode={518172}, 
                country={China}}
                
\affiliation[3]{organization={Shenzhen University},
            city={Shenzhen},
            postcode={518037}, 
            country={China}}
\affiliation[4]{organization={Eye Center, The Second Affiliated Hospital, School of Medicine, Zhejiang University, Zhejiang Provincial Key Laboratory of Ophthalmology},
            city={Hangzhou},
            postcode={310000}, 
            country={China}}    

\fntext[fn1]{These authors contributed equally to this work.}

\cortext[cor1]{Corresponding author}

\begin{abstract}
Ultra-Wide-Field Fluorescein Angiography (UWF-FA) enables precise identification of ocular diseases using sodium fluorescein, which can be potentially harmful. Existing research has developed methods to generate UWF-FA from Ultra-Wide-Field Scanning Laser Ophthalmoscopy (UWF-SLO) to reduce the adverse reactions associated with injections. However, these methods have been less effective in producing high-quality late-phase UWF-FA, particularly in lesion areas and fine details. Two primary challenges hinder the generation of high-quality late-phase UWF-FA: the scarcity of paired UWF-SLO and early/late-phase UWF-FA datasets, and the need for realistic generation at lesion sites and potential blood leakage regions. This study introduces an improved latent diffusion model framework to generate high-quality late-phase UWF-FA from limited paired UWF images. To address the challenges as mentioned earlier, our approach employs a module utilizing Cross-temporal Regional Difference Loss, which encourages the model to focus on the differences between early and late phases.
Additionally, we introduce a low-frequency enhanced noise strategy in the diffusion forward process to improve the realism of medical images. To further enhance the mapping capability of the variational autoencoder module, especially with limited datasets, we implement a Gated Convolutional Encoder to extract additional information from conditional images. Our Latent Diffusion Model for Ultra-Wide-Field Late-Phase Fluorescein Angiography (LPUWF-LDM) effectively reconstructs fine details in late-phase UWF-FA and achieves state-of-the-art results compared to other existing methods when working with limited datasets. Our source code is available at: \href{https://github.com/Tinysqua/LPUWF-LDM}{https://github.com/Tinysqua/LPUWF-LDM}.
\end{abstract}

\begin{graphicalabstract}
\includegraphics[width=0.7\textwidth]{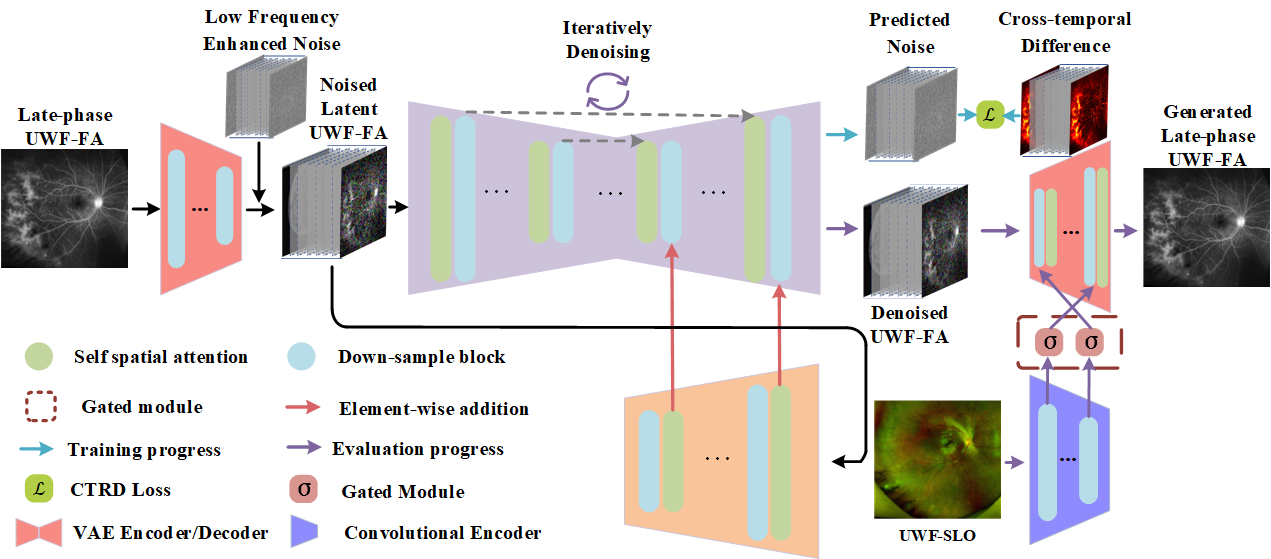}
\end{graphicalabstract}

\begin{highlights}
\item A cross-modal latent diffusion model focused on generating late-phase UWF-FA. 
\item Enhanced Variational Autoencoder architecture for cross-modal generation on limited data. 
\item A more suitable noise strategy for ophthalmic image distribution in diffusion forward process. 
\item A loss function that promotes the diffusion model to focus on lesion areas in an unsupervised manner.
\end{highlights}

\begin{keywords}
Diffusion Model \sep Loss Enhancement \sep Cross-modal Generation \sep Ultra-wide-field Fundus Photo \sep Variational Autoencoder
\end{keywords}

\maketitle

\section{Introduction}

Ultrawide field fluorescein angiography (UWF-FA) is a dynamic imaging technique for diagnosing and treating fundus-related diseases \citep{ref_1, ref_4, ref_21}. This procedure involves injecting a dye into a patient's vein, which travels to the eye's fundus, potentially causing nausea or vomiting. It can also pose risks for patients with serious heart conditions. In contrast, Ultra-Wide-Field Scanning Laser Ophthalmoscopy (UWF-SLO) rapidly scans the retina using laser imaging and does not adversely affect the patient. However, UWF-SLO produces less detailed vascular images. Previous studies, such as VTGAN \citep{kamran2021vtgan} and Reg-GAN \citep{rezagholiradeh2018reg}, have attempted to address this issue by transforming UWF-SLO images into UWF-FA images using cross-modal generation methods. These approaches, however, only supervise learning with a single UWF-SLO and an early-phase UWF-FA, neglecting the dynamic nature of UWF-FA. Issues related to retinal structure, like central serous chorioretinopathy \citep{chen2021automatic}, are often assessed in the late phase. The scarcity of paired UWF-SLO, early-phase, and late-phase UWF-FA datasets makes generating high-quality UWF-FA from UWF-SLO a significant challenge. Two primary technical difficulties arise: (1) producing high-quality UWF-FA with detailed lesion information to aid diagnosis, and (2) maintaining the quality of generated late-phase UWF-FA despite limited paired datasets.

\begin{figure*}[t!]
    \centering
    \includegraphics[width=1\textwidth,height=0.28\textheight]{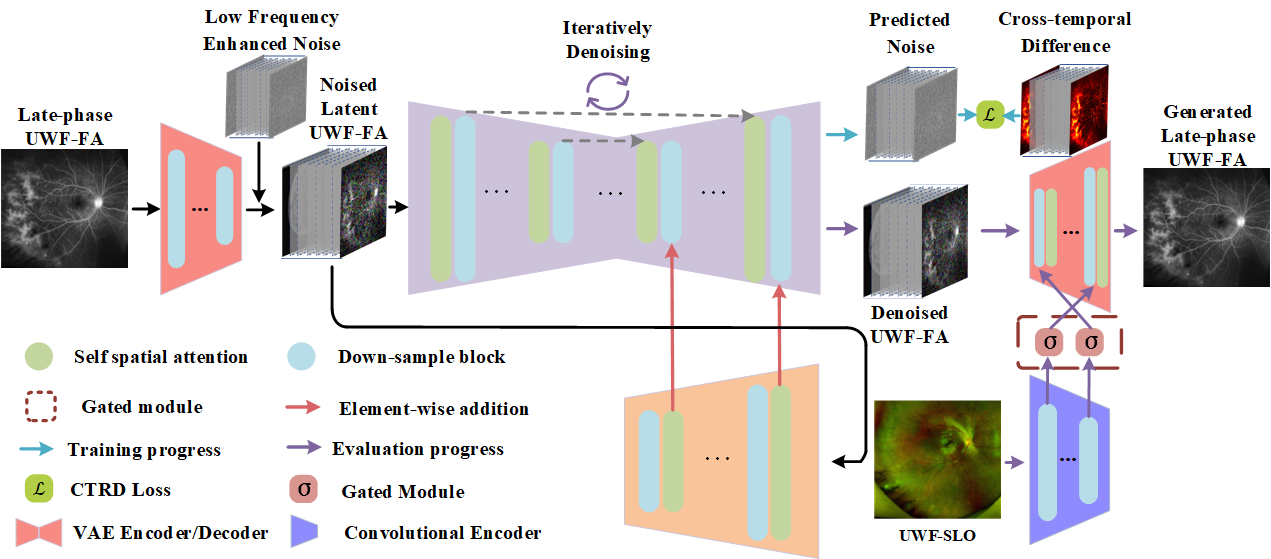}
    \caption{Overall architecture of LPUWF-LDM. It encompasses a VAE module with a Gated Convolutional Encoder, a noise addition module utilizing low-frequency enhanced noise, a conditional encoder module for input conditional images, and a backbone for noise prediction trained via CTRD Loss.}
    \label{fig:Overall_architecture}
\end{figure*}

To address these challenges, we propose an enhanced latent diffusion framework. This framework utilizes a diffusion network and a Variational Autoencoder (VAE) \citep{kingma2013auto} to generate late-phase UWF-FA images with high fidelity and stability. Besides, unlike traditional diffusion models that generally require extensive training on large datasets and are typically optimized for natural images, our approach enhances the traditional latent diffusion framework by improving its performance with smaller datasets and increasing its sensitivity to the differences between early and late-phase UWF-FA images. This makes our model more suitable for training on medical retina data, ensuring high-quality image generation even with limited data availability.

In this paper, we begin by training our model on a multicenter dataset of UWF-SLO and early-phase UWF-FA images to capture structural information. Following this, we train the model on a smaller dataset of paired UWF-SLO and late-phase UWF-FA images. To enhance the model's focus on temporal discrepancies, we incorporate a Cross-temporal Regional Difference Loss (CTRD Loss) into the loss function. Given that ophthalmic images are rich in low-frequency components and traditional noise addition methods often vary in their handling of high and low frequencies, we employ a low-frequency enhanced noise strategy. This approach improves the model's ability to infer medical images that are abundant in low-frequency information. For the VAE component, we introduce a Gated Convolutional Encoder. This encoder extracts additional information from UWF-SLO images, assisting in pixel-space reconstruction by filtering useful information through the gated module. Additionally, we propose a preprocessing method for UWF fundus photographs, which includes sharpening UWF-SLO images and aligning early and late-phase UWF-FA images.

To demonstrate the effectiveness of our LPUWF-LDM framework, we benchmarked it against leading contemporary image generation models using our proprietary UWF image datasets. We employed both qualitative and quantitative metrics, including the Fréchet Inception Distance (FID) \citep{binkowski2018demystifying}, Inception Score (IS) \citep{chong2020effectively}, Peak Signal-to-Noise Ratio (PSNR) \citep{sara2019image}, and Multi-Scale Structural Similarity Index (MS-SSIM) \citep{wang2004image}.

Our contributions can be summarized as follows:
\begin{itemize}
    \item 	To the best of our knowledge, this is the first study to train and evaluate diffusion models for generating late-phase UWF-FA images from UWF-SLO, eliminating the need for dye injections. We provide pairs of early-phase and late-phase UWF-FA images that have undergone registration and noise reduction.
\item To improve performance on small datasets, we implemented CTRD Loss and incorporated a Gated Convolutional Encoder for VAE, enhancing the original AutoencoderKL. We also employed low-frequency enhanced noise to better suit the diffusion process for ophthalmic medical images.
\item 	Through extensive experiments and comparisons, we demonstrated that the proposed LPUWF-LDM achieves state-of-the-art performance on clinical proprietary UWF image datasets.
\end{itemize}
\begin{figure}[t!]
    \centering
    \includegraphics[width=0.7\textwidth]{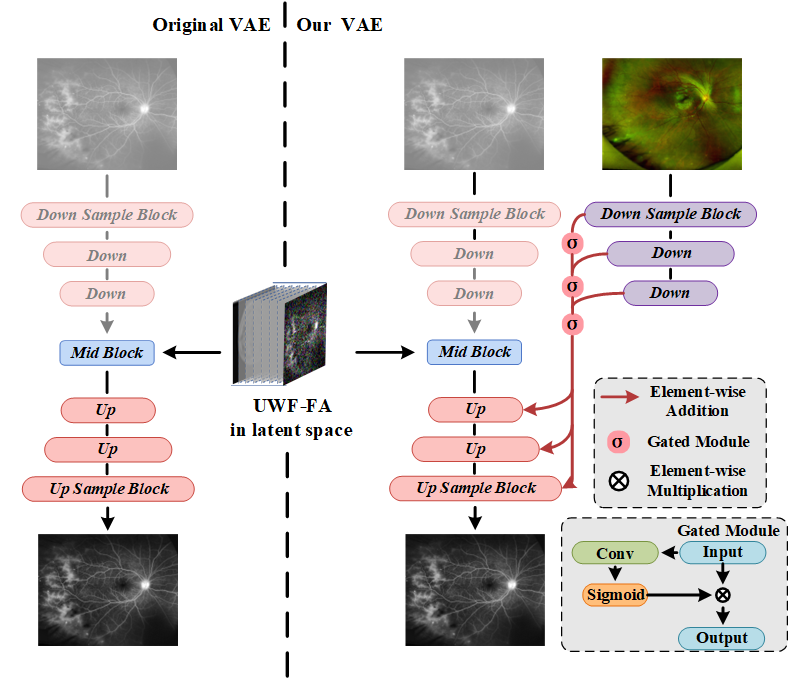}
    \caption{Details of the Gated Convolutional Encoder for the VAE framework. A comparison between the traditional VAE approach and our VAE method. Our method augments the original VAE framework with a Gated Convolutional Encoder, which consists of a Downsample Module and a Gated Module.}
    \label{fig:Conditional-vae}
\end{figure}

\section{Related Work}
\subsection{Cross-modal Generation}

In medical imaging, various modalities such as magnetic resonance imaging, positron emission tomography, UWF-SLO, and UWF-FA are commonly used for patient examinations. Establishing nonlinear mappings between these modalities using deep learning can provide multimodal information that aids in diagnosis while avoiding the side effects associated with certain imaging procedures. Cross-modal image generation, particularly generating late-phase UWF-FA from UWF-SLO, requires high-quality output characterized by accurate structural and pathological information. The U-Net architecture, which utilizes convolutional neural networks for upsampling and downsampling and incorporates skip connections, was initially employed for cross-modal generation \citep{li2019ct, dovletov2022grad}. Generative adversarial networks (GANs) \citep{goodfellow2020generative} and their variant, conditional GAN \citep{vaidya2022perceptually, uzunova2020memory}, built on the U-Net architecture, use a discriminator to judge the authenticity of generated content. This adversarial approach improves the quality of cross-modal generated images by reducing local blurriness.

In ophthalmology, Kamran et al. \citep{kamran2021vtgan} and Fang et al. \citep{fang2023uwat} have successfully achieved realistic cross-modal generation on standard and UWF fundus images. However, GAN-based methods often face issues such as local distortions and mode collapse, making them unstable for practical use. Recently, denoising diffusion probabilistic models have emerged as a more stable alternative. These models gradually introduce noise to the images, converting them into pure noise, and then train the model to map them back to the original state, generating high-fidelity images. Due to these advantages, diffusion models have been applied to medical image generation \citep{pan2023cycle, song2023deep}.

Despite their promise, diffusion models still struggle with poor performance on small datasets and a lack of focus on lesion content. To address this, our approach enhances the learning of the diffusion model in lesion areas through CTRD Loss, allowing it to generate finer lesion details even on smaller datasets.

\begin{figure}[t!]
    \centering
    \includegraphics[width=0.7\linewidth]{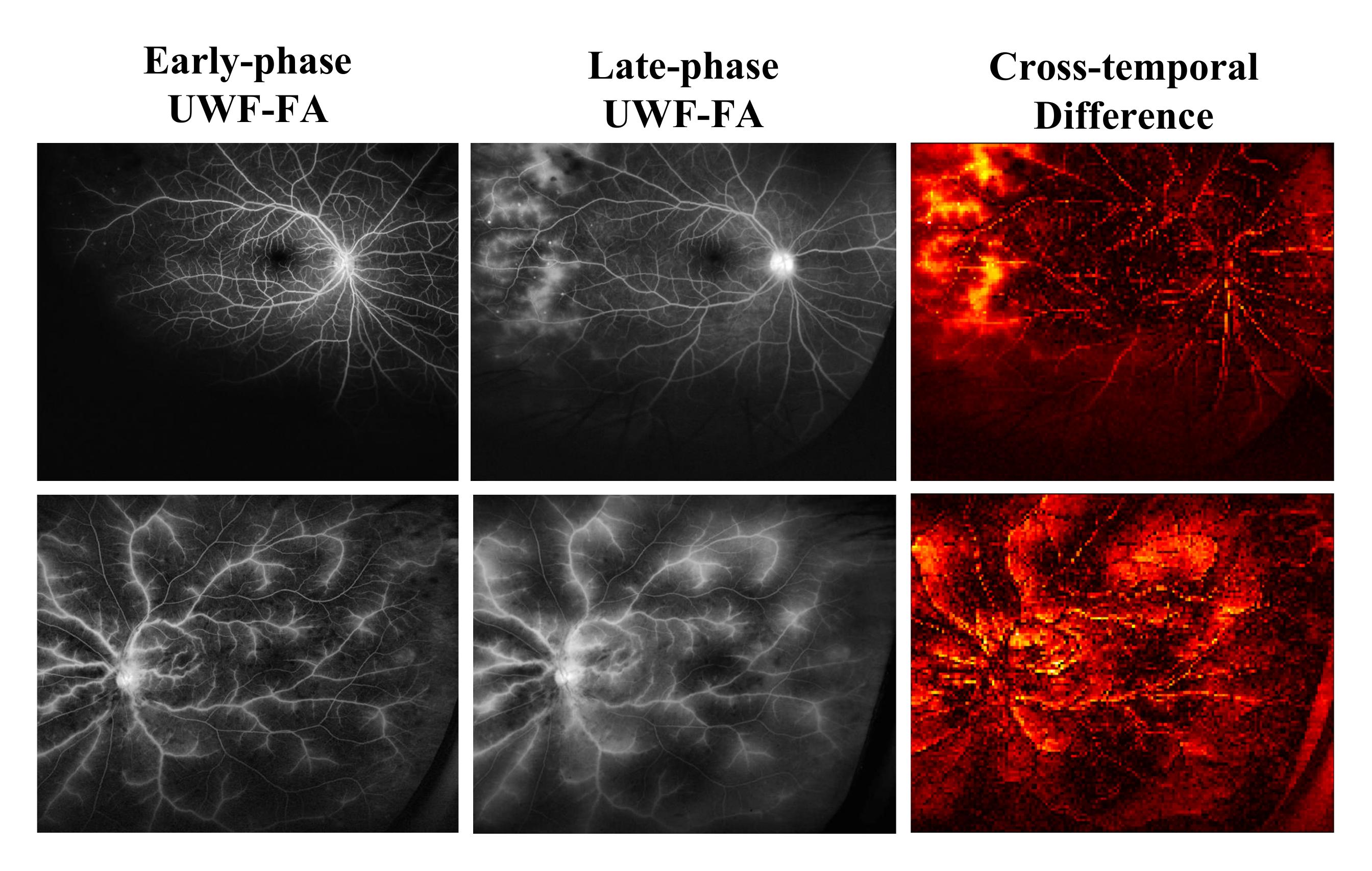}
    \caption{Two pairs of early-phase and late-phase UWF-FA images and their Cross-temporal Regional Differences. The left column shows early-phase UWF-FA, the middle column displays late-phase UWF-FA, and the rightmost column presents the Cross-temporal Differences.}
    \label{fig:Difference}
\end{figure}

\subsection{Diffusion Models combined with VAE}

Traditional diffusion models, while stable and capable of producing high-quality images, require substantial computational resources. Recent models, such as Latent Diffusion Model (LDM) and Stable Diffusion, use a VAE to downsample images into a latent space, making them effective for generating high-resolution ophthalmic images \citep{jang2023taupetgen, zhu2023make}. However, LDMs are not directly suitable for cross-modality generation due to their primary reliance on text as a conditioning input. 

ControlNet \citep{zhang2023adding} introduced the potential for high-resolution and high-quality generation in late-phase UWF-FA tasks by using the source image as a condition. Because of these advantages, methods based on controlled and LDM have begun to be practiced on medical images \citep{go2024generation, kim2024feasibility}. Despite this advancement, LDMs still face challenges with poor performance on small datasets, and the incorporation of VAEs introduces additional issues. LDMs typically employ a VQGAN, which combines vector quantization \citep{rombach2022high} with a GAN discriminator, or an AutoencoderKL that includes a traditional VAE with an added discriminator. VAEs need to be trained on a wide range of datasets; otherwise, even if the diffusion part is well-fitted, overall image quality can degrade due to a decline in VAE-generated quality. Moreover, the effectiveness of converting pathologies in cross-modality generation remains limited.

To address these issues, we have enhanced the original VAE with a Gated Convolutional Encoder that leverages transfer learning. This enhancement enables the selective use of UWF-SLO condition information during the reconstruction of late-phase UWF-FA images. Additionally, sharpening the UWF-SLO images allows for more effective conversion of pathologies, thereby improving the quality and accuracy of cross-modality generation.

\begin{figure}[t!]
    \centering
    \includegraphics[width=0.6\linewidth]{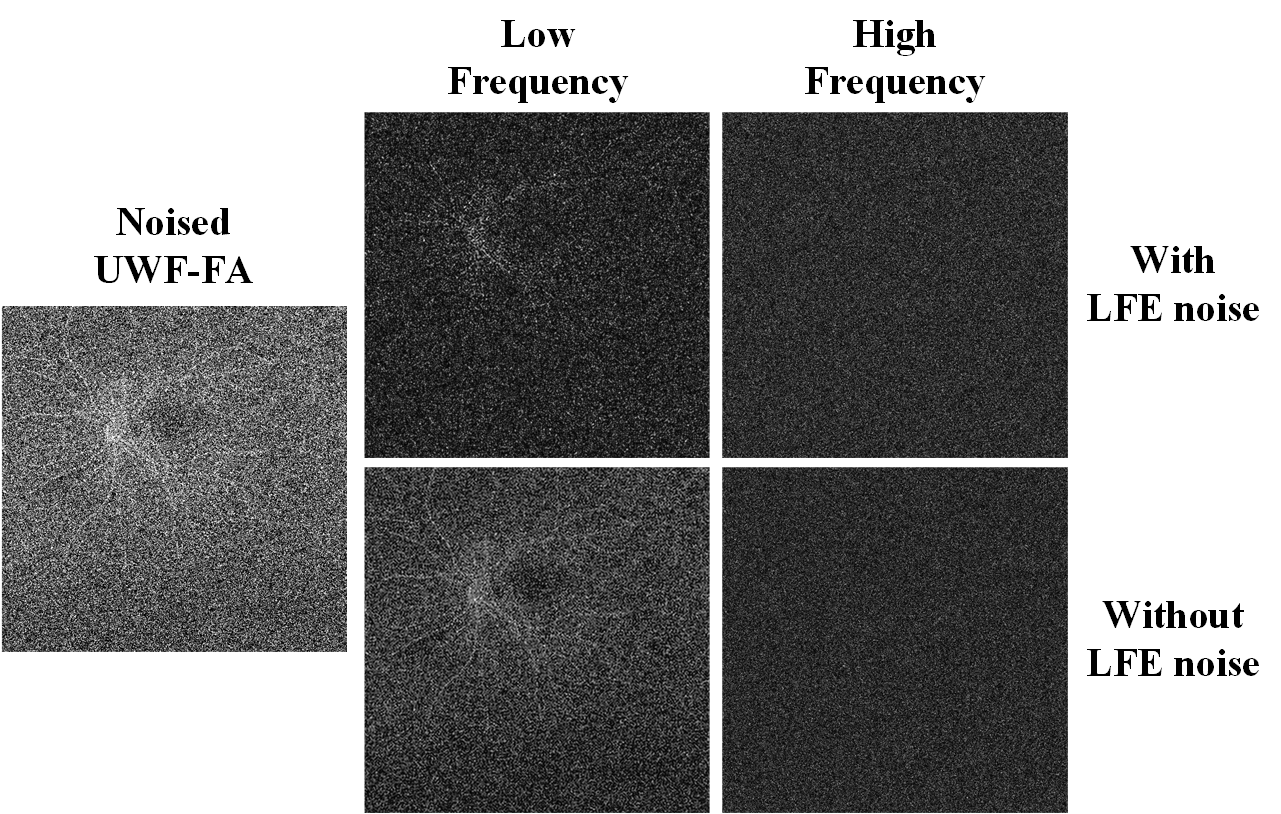}
    \caption{High and low-frequency division of noised UWF-FA using fourier transform. The top row shows the visualization of high and low frequencies with Low-Frequency Enhanced Noise applied, while the bottom row displays those without it.}
    \label{fig:Low Frequency Enhanced noise}
\end{figure}

\section{Methods}
The overall architecture of our method, as illustrated in Fig. \ref{fig:Overall_architecture}, comprises four principal components: a VAE module for upsampling and downsampling, a diffusion forward noise addition module, a backbone module using a U-Net architecture for noise prediction, and a Control Encoder module for spatial conditioning.

Given a dataset ($x^i$, $y_E^i$, $y^i$)~D, where $x^i$ denotes UWF-SLO, $y_E^i$ represents early-phase UWF-FA, and $y^i$ signifies late-phase UWF-FA, the training process is as follows: Initially, the model is trained on pairs of $x^i$ and $y_E^i$ to learn structural information. Subsequently, it is trained on pairs of $x^i$ and $y^i$. During inference, the model generates the corresponding $y^i$ based on the input image $x^i$.

For the diffusion component, $y^i$ is first compressed into the latent space $y_L^i$ via the VAE's Encoder. The forward noise addition module then incorporates our proposed Low-Frequency Enhanced Noise, transforming $y_L^i$ into a noise-augmented image at step t, denoted as $y_t^i$. Both $x^i$ and $y_t^i$ are concurrently input into the Control Encoder, which extracts features and feeds them into the backbone network to predict the noise added to $y_L^i$, represented as $\tilde{\epsilon}$. Training involves augmenting the original diffusion MAE loss with a CTRD Loss. During inference, what was originally input into the backbone and Control Encoder as $y_L^i$ turns into pure noise $\epsilon$. This noise is then iteratively refined back to $y_L^i$ through a recursive process. Finally, the VAE's Decoder reverts $y_L^i$ to $y^i$. During the VAE Decoder's operation, the Gated Convolutional Encoder takes $x^i$ as input, and the intermediate feature maps it produces, after passing through the gating mechanism, are input into the Decoder to aid in the restoration process of $y^i$.

\subsection{Gated Convolutional Encoder For VAE}
Our cross-modal tasks use images as conditions, which will provide rich spatial information, such as the details of the blood vessels. Notably, the addition of this supplementary information has been observed to improve the generation quality and generalization performance of VAEs, especially when applied to smaller datasets. However, UWF-SLO images, which serve as additional information, often contain significant noise, such as orbital areas around the eyes. To address this, we propose a gating mechanism to filter out noise in the conditional images. 

The detailed architecture is presented in Fig. \ref{fig:Conditional-vae}, consisting of VAE backbone and Gated  Convolutional Encoder. The VAE backbone contains an Encoder ($E_{FA}$) and a Decoder ($D_{FA}$), along with a discriminator to enhance the quality of the generated images. Initially, the late-phase UWF-FA image ($y^i$) is fed into $E_{FA}$. It first passes through a convolutional layer with a kernel size of 3, a stride of 1, and padding of 1 (Conv, k=3, s=1, p=1). Next, it moves through three Down blocks, each comprising two Residual blocks and one Down-sample block. The output of $E_{FA}$ provides the mean and variance in the latent space, from which the latent vector ($y^i_L$) is sampled. The latent vector ($y^i_L$) is then fed into $D_{FA}$, which consists of three Up blocks. Since the Decoder also receives additional spatial information, each Up block checks for this extra information and incorporates it into the current layer's vector if available. The vector then passes through two Residual blocks and a self-attention block. Finally, it moves through a Group Normalization layer and another convolutional layer (Conv, k=3, s=1, p=1) to reconstruct $y^i_L$ into the final output.

The Gated Convolutional Encoder consists of three Gated Down blocks, each incorporating a convolutional layer with a kernel size of 3, stride of 2, and padding of 1 to compress the input image $x^i$. Additionally, each Down block includes a Gate Module, which features a convolutional layer (Conv, k=3, s=1, p=1) followed by a non-linear Sigmoid layer. The relationship between the Down blocks is depicted in Fig. \ref{fig:Conditional-vae}. For an input $y^i_n$ at layer n, the next layer's input $y_{n+1}^i$ is computed as:
\begin{flalign}&&
y_{n+1}^i  = \sigma \left ( gate\underline{~}module \left ( Conv \left (  y^i_n\right )    \right )    \bullet Conv\left ( y_n^i \right )\right ),&&
\end{flalign}
where $\sigma$ denotes the non-linear computation, Conv represents the convolution operation and $\bullet$ represents element-wise multiplication.

Training the entire VAE includes a two-step process. First, we perform transfer learning from the AutoencoderKL of Stable Diffusion and then fine-tune it on our dataset. The loss function used is:

\begin{flalign}
    \label{equ:G_loss}&&
    \mathbb{L}=\left \| y^i-D_{FA}\left ( E_{FA}\left ( y^i \right )  \right )  \right \| ^2_2+\log \left ( 1-D\left ( D_{FA}\left ( E_{FA}\left ( y^i \right )  \right )  \right )  \right ) + \notag   &&\\
    &&Perc\left ( y^i,D_{FA}\left (E_{FA}\right )\right )+KL\left ( E_{FA}\left ( y^i \right ) \right ), &&  
\end{flalign}
where $D$ denotes the discriminator, Perc represents the Perception loss, and KL denotes the Kullback-Leibler loss. The loss function for the discriminator $D$ is:
\begin{flalign}&&
\mathbb{L}\left (D\right )=\log (1-D(y^i)) +\log(D(D_{FA}(E_{FA}(y^i)))).&&
\end{flalign}
When training the Gated Convolutional Encoder, we set only its parameters to be trainable and remove the KL loss from the loss function in Eq. \ref{equ:G_loss}, while keeping the other components unchanged. Training continues until the model is well-fitted.

\begin{figure}[t!]
    \centering
    \includegraphics[width=0.8\linewidth]{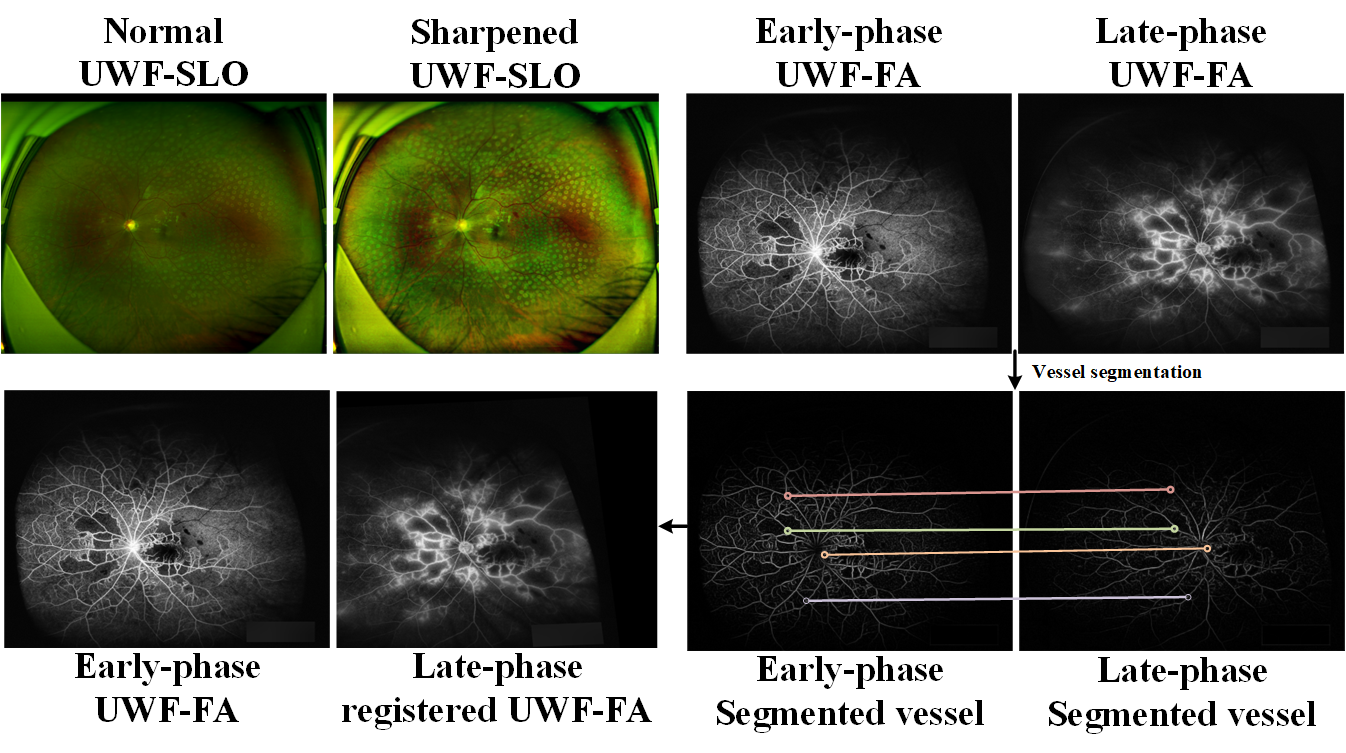}
     \caption{The Effects of UWF photos preprocessing. The top left corner demonstrates the effect of image sharpening, and the bottom right corner demonstrates the process of UWF-FA registration.}
    \label{preprocessing}
\end{figure}

\subsection{Cross-temporal Reginal Difference Loss}
The early-phase UWF-FA is very clear in terms of structural information, such as the position of arteries, veins and optic disc. Therefore, it is better for the model to learn the mapping from UWF-SLO to early-phase UWF-FA efficiently. However, while the structural information can guide the realism of image generation, it is disadvantageous in generating detailed information related to lesions, as more lesion information is actually reflected in the late-phase UWF-FA. Therefore, the module that is designed aims to solve two problems: 1) How to locate the lesion areas without the need for manual annotation. 2) How to make the diffusion model focus more on these areas. 
Therefore, we propose the Cross-temporal Regional Difference Loss. Specifically, we first register the early-phase and late-phase UWF-FA images, then compute the pixel-wise absolute difference, and normalize the result to the range (0, 1) to obtain a heatmap of the same shape and size as the UWF-FA images. Regions with higher values indicate greater differences between the early and late phases, while lower values correspond to smaller differences. From Fig. \ref{fig:Difference}, we can observe that the uninformative black background has lower values, while the vascular regions and the leakage area on the left exhibit higher values. Statistically, the average difference induced by fluorescence leakage is 0.87, while that of neovascularization is 0.71. The higher-valued locations signify that the model should focus more on generating these regions, forming an unsupervised attention to the lesion areas. Since the diffusion is trained in the latent space, this heatmap is resized to match the same size as the latent space. Therefore, the overall learning objective based on diffusion model is: 
\begin{flalign}
&&
\mathbb{L}=\lim_{E_{FA}(y^i_0),t,x^i,\epsilon \sim N(0,1)} \left [{ \alpha\cdot \omega (y^i_E,t,x^i)}\right.\left.{\cdot \left \| \epsilon -\epsilon _\theta (y^i_t,t,x^i) \right \| ^2_2+\left \| \epsilon -\epsilon _\theta(y^i_t,t,x^i)  \right \|^2_2} \right ],&&
\end{flalign}
the $\alpha$ is a hyperparameter that represents the weight of this loss term. In the experiments, $\alpha$ gradually increases from 0.25 to 1 
during the first half of the training epochs, and remains at 1 for the second half of the training epochs. The $w$ denotes the aforementioned heatmap, which is related to the early-phase and late-phase UWF-FA. $y_t^i$ represents the noisy image, where $t$ indicates the current step of adding noise, and $x^i$ is the conditional input UWF-SLO. 

\begin{figure*}[t]
    \centering
    \includegraphics[width=1\linewidth]{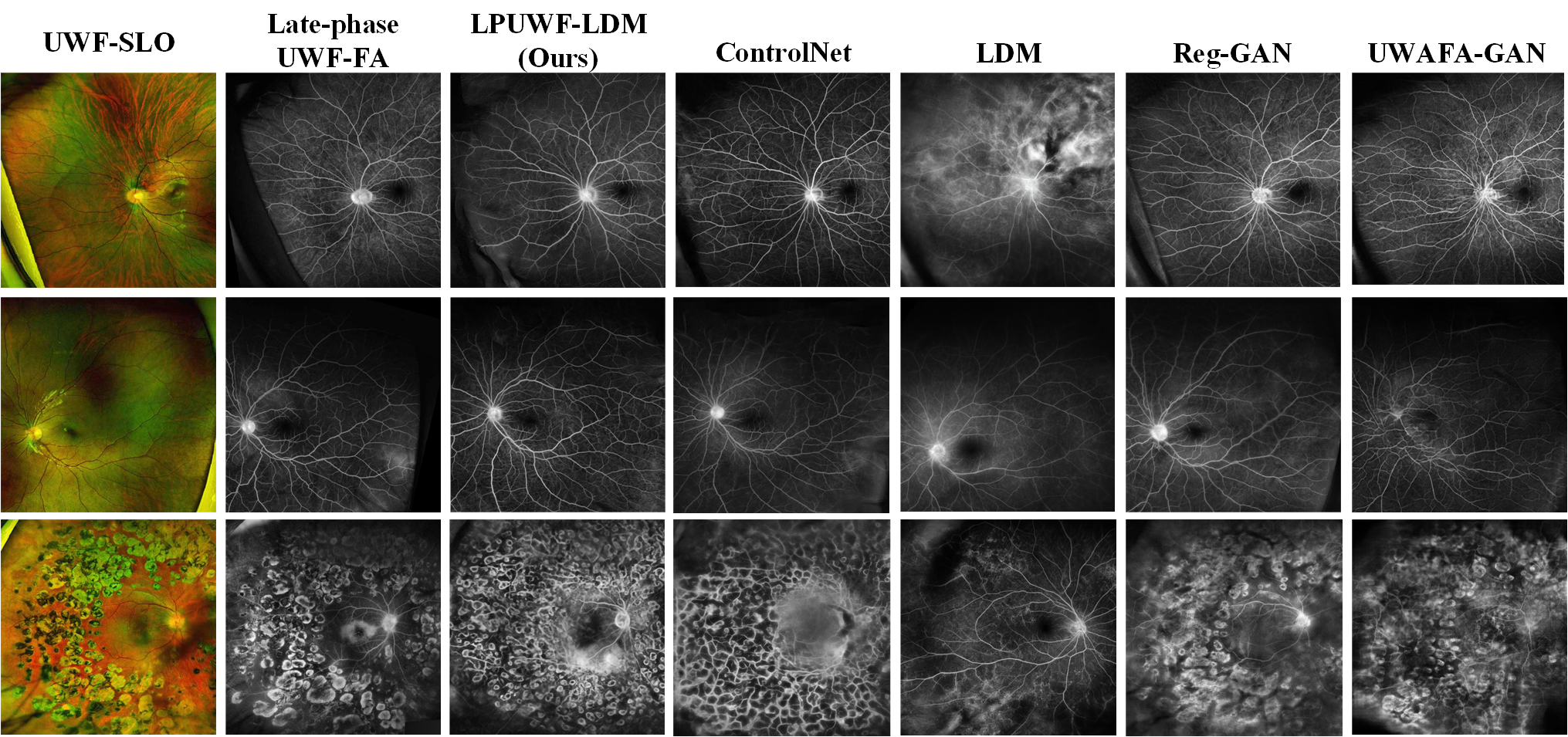}
     \caption{Comparison of generating late-phase UWF-FA between different generation networks.}
    \label{fig:qualitative_com}
\end{figure*}

\subsection{Low Frequency Enhanced Noise}
The traditional strategy of injecting noise into images by adding identically distributed (i.i.d.) Gaussian noise affects high-frequency and low-frequency information differently. When a noisy UWF-FA image is decomposed into these components, a noticeable disparity in noise injection between high-frequency and low-frequency regions is observed. Specifically, low-frequency regions experience less noise interference compared to high-frequency regions. This discrepancy can impair the model's ability to restore low-frequency details, which are abundant in medical images.

To address this issue, we aim to increase the noise interference on low-frequency information. We achieve this by adding a low-frequency noise component to the original Gaussian noise, which follows a $N(0,1)$ distribution. To generate this low-frequency noise, we first sample a value $a$ from $N(0,1)$ and a scale factor $\beta$ from $N(0,0.5)$ to control the degree of low-frequency noise. We then multiply $a$ and $\beta$, and replicate the resultant product $a\beta$ across the height and width dimensions of the image, creating a low-frequency noise term $\epsilon_L$ with no numerical variation. Fig. \ref{fig:Low Frequency Enhanced noise} shows the separated images of high and low frequencies after increasing the low frequency. Under the ordinary noise addition strategy, the low frequency still maintains relatively clear textures, while our strategy makes the damage of high- and low-frequency noise more balanced.

By adding this low-frequency noise term to the original noise, the new noise vector $y_t$ maintains the i.i.d. assumption. Therefore, the new $y_t$ can be represented as follows:
\begin{flalign}&&
y^i_t=\sqrt{\overline{\alpha}_t } y^i_0+\sqrt{1-\overline{\alpha}_t } \left ( \epsilon +\epsilon_L  \right ), &&
\end{flalign}
where $\epsilon$ and $\epsilon_L$ denote pure noise and low-frequency noise, respectively.

\subsection{Preprocessing of UWF photos}
\label{section:preporcessing}
The raw UWF fundus images do not perform well in direct training due to two significant sources of noise: 1) the blood vessels in the UWF-SLO are mixed with the laser background, and 2) there is misalignment between the early-phase and late-phase UWF-FA. To address these issues, we have incorporated image sharpening for UWF-SLO and registration between early-phase and late-phase UWF-FA into our pipeline.

To enhance the UWF-SLO images, we apply histogram equalization techniques separately to the RGB channels before merging them back into the UWF-SLO image. This process makes the vessel colors more distinguishable from the background, as illustrated in Fig. \ref{preprocessing}. For registering the early-phase and late-phase UWF-FA images, we use the SIFT algorithm to detect keypoints and compute descriptors. The best matching points are then used to calculate the homography matrix and perform a perspective transformation, aligning the late-phase UWF-FA to the early-phase.

However, due to significant choroidal background fluorescence noise, the registration algorithm struggles to detect enough keypoints for matching. Therefore, it is crucial to minimize the noise in the input images. To achieve this, we use a multi-scale linear filter for vessel segmentation. The filter formula is as follows:

\begin{figure*}[t]
    \centering
    \includegraphics[width=0.85\linewidth]{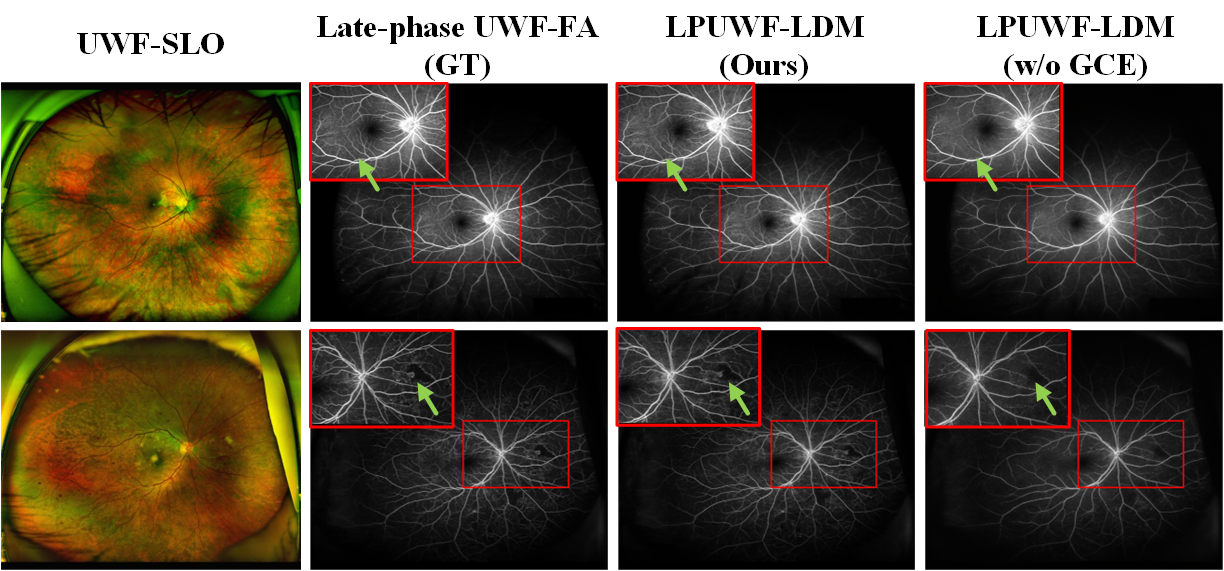}
     \caption{The reconstruction effect of VAE with or without Gated Convolutional Encoder. The green arrows indicate the location of capillaries and the quality of reconstruction.}
    \label{fig:w/o_GCE}
\end{figure*}
\begin{flalign}&&
I_{\alpha} = \frac{\sum_{N} (I_{m} - \alpha \cdot (\bar{G}_x + \bar{G}_y) - I) + I}{N+1},&& \end{flalign}
where \( I \) is the grayscale value of the window center, \( I_m \) is the maximum average grayscale value of the fan-shaped region in the window, \( \overline{G}_x \) and \( \overline{G}_y \) are the average gradient values in the horizontal and vertical directions of the fan-shaped region, \( N \) is the number of fan-shaped regions in the window, and \( \alpha \) is an adjustable parameter that controls the influence of gradient compensation. In this filtering process, a window is defined, and \( N \) sectors with variable radii are drawn from the center. The average grayscale value for each sector is computed, incorporating compensation based on local average gradient values. The data obtained at different scales are linearly combined to calculate the multi-scale filtered grayscale value \( I_\alpha \) at the center of the window.

After this filtering process, the vessels are effectively segmented from the background, significantly enhancing the image registration effect. Fig. \ref{preprocessing} depicts the UWF-FA before registration, the segmented blood vessel image during registration, and the final result after registration.

\section{Experiments}
\subsection{Dataset}
We utilized datasets from two collaborating hospitals. The first hospital's dataset primarily includes paired UWF-SLO images and early-phase UWF-FA images. These data underwent rigorous screening to exclude images that could compromise clinical diagnosis, such as those with capture intervals exceeding three months, visible fresh bleeding, severe eyelash occlusion, or poor focus. Ultimately, 304 pairs of high-quality images were selected from this hospital. After data augmentation, these images were fully utilized for extensive pre-training of the model. The second hospital's dataset contains paired UWF-SLO images, early-phase UWF-FA images, and late-phase UWF-FA images. After filtering, we identified 387 image pairs for our dataset, with each set containing all three image types. We then divided these images into 309 pairs for the training set and 78 pairs for the test set. The classification was based on physician definitions, with early-phase designated as 0 to 15 seconds and late-phase as after 10 minutes.

To ensure data consistency, we adjusted the image resolution to a uniform 853x682 pixels across both datasets and applied various enhancement measures. These included random rotation within \([-5^\circ, 5^\circ]\), cropping to 512x512 pixels, and random flipping. These enhancements effectively expanded the original training dataset from 304 pairs to 19,456 pairs and the 309 sets to 6,192 sets, totaling 25,648 UWF-SLO images, 25,648 early-phase UWF-FA images, and 6,192 late-phase UWF-FA images.
\subsection{Evaluation Metrics}
We adopted four evaluation metrics: FID ($\downarrow$), IS ($\uparrow$), PSNR ($\uparrow$), and MS-SSIM ($\uparrow$), which together provided a comprehensive framework for assessing the quality of generated images. FID is a crucial metric for evaluating the fidelity and realism of generated images, as it captures perceptual differences beyond mere pixel-level disparities. IS measures the diversity and quality of generated images through the softmax probability distribution within the Inception model, with a higher IS value indicating clearer and richer image content. PSNR quantifies distortion by calculating the mean squared error between images, directly reflecting the degree of quality loss or degradation in generated images compared to their real counterparts. MS-SSIM, on the other hand, offers a detailed assessment of structural similarities between images from a multi-scale perspective, encompassing dimensions such as luminance, contrast, and structural information. All evaluations based on these metrics were conducted on the test set of the second dataset, which comprised 309 training sets and 78 test sets.

\subsection{Implementation Details}
In this study, we employed the Pytorch 2.0 cuda 11.8 framework to construct the LPUWF-LDM network and trained it using two NVIDIA A100 GPUs. The VAE and diffusion components were each trained for 1000 epochs. The VAE component, trained with a batch size of 8, required 52 hours. While the diffusion component, with a batch size of 32, took 33 hours. The training process began with the VAE itself, followed by freezing the VAE to train the Gated Convolutional Encoder. Subsequently, both the VAE and the Gated Convolutional Encoder were frozen. The diffusion model was then initially trained on paired UWF-SLO and early-phase UWF-FA data, and later on paired UWF-SLO and late-phase UWF-FA data. Each component inherited parameters from Stable Diffusion and underwent transfer learning. Both models were optimized using the Adam algorithm, with a learning rate and betas set to (0.9, 0.999).
\begin{table}[t]
    \caption{Comparison of generative metrics for various generation networks. The best and second-best performances are indicated in red and blue colors, respectively.}
    \centering
    \begin{tabular}{llllllllll}\toprule
        ~ & FID($\downarrow$) & IS($\uparrow$) & PSNR($\uparrow$) & MS-SSIM($\uparrow$)   \\  \hline
        ControlNet \citep{zhang2023adding} & 94.6701 & 1.5171 & 28.6113 & \textcolor{blue}{0.6749}\\ 
        LDM \citep{rombach2022high} & 97.6410 & 1.6055 & 28.3523 & 0.2859 \\ 
        Stable Diffusion \citep{podell2023sdxl} & 97.4531 & 1.6600
        & 29.1842 & 0.6656 \\
        Pix2pixHD \citep{wang2018high} & 95.5519 & 1.5389 & 28.0225 & 0.4031 \\
        Reg-GAN \citep{rezagholiradeh2018reg} & \textcolor{blue}{79.7899} & \textcolor{red}{\textbf{1.7726}} & \textcolor{blue}{28.6854} & 0.6358 \\ 
        UWAFA-GAN \citep{ge2024uwafa} & 85.8738 & 1.6571 & 28.0934 & 0.6653   \\ 
        \textbf{LPUWF-LDM(Ours)} & \textcolor{red}{\textbf{77.6596}} & \textcolor{blue}{1.7578} & \textcolor{red}{\textbf{30.6727}} & \textcolor{red}{\textbf{0.7104}}\\ 
          \hline
    \end{tabular}
    \label{Tab:comparison}
\end{table}
\subsection{Comparison}
In this experiment, we compared our method against various other generative models, selecting three methods each from GAN networks and diffusion networks for comparison. For the GAN network methods, we chose Pix2pixHD \cite{wang2018high}, Reg-GAN and UWAFA-GAN. Pix2pixHD can generate high-quality images at a resolution of 2048x1024 through its multi-scale generator design and the incorporation of perceptual loss. Reg-GAN combines CycleGAN with a registration module, enabling it to generate high-quality images even with slightly misaligned ophthalmic image datasets. UWAFA-GAN, on the other hand, is the first method to use GANs to create early-phase UWF-FA from UWF-SLO, achieving promising results.

For diffusion methods, we selected Stable Diffusion \citep{podell2023sdxl}, LDM and ControlNet. Stable Diffusion and LDM compress images into latent space and train on it, which allows them to generate large-sized medical images. ControlNet allows spatial information constraints on generation. For each model's training, we followed the default hyperparameters given in their respective codes. We pre-trained each model on paired UWF-SLO and early-phase UWF-FA data, then fine-tuned them on paired UWF-SLO and late-phase UWF-FA data.

As shown in Table \ref{Tab:comparison}, our method outperforms others in terms of FID, PSNR, and MS-SSIM, and also achieves competitive results in IS. Compared to Pix2pixHD, our method reduces the FID by 17.8923, and increases IS, PSNR and MS-SSIM by 0.2189, 2.6502, 0.3073. Compared to UWAFA-GAN, our method reduces the FID by 8.2142, while increasing IS, PSNR, and MS-SSIM by 0.1007, 2.5793, and 0.0451, respectively. This indicates that our diffusion-based method generates higher-quality images with lower noise levels. Compared to Reg-GAN, our model reduces FID by 2.1303 and improves PSNR and MS-SSIM by 1.9873 and 0.0746, respectively, though it has a slightly lower IS of 0.0148, which might be due to adversarial training in Reg-GAN that produces images with higher clarity. 

Despite both being diffusion models, our method surpasses Stable Diffusion, ControlNet and LDM in IS, PSNR, and MS-SSIM, and significantly outperforms them in FID by 19.7935, 17.0105 and 19.9814, respectively. This demonstrates that our method maintains excellent generalization even on a limited dataset. Figure \ref{fig:qualitative_com} presents a qualitative comparison of our method with other competitive generative methods. The first two rows show the normal transition from UWF-SLO to late-phase UWF-FA, while the last row illustrates the transition in the presence of obvious lesions. Our model excels in generating vascular and disc structures, with more continuous vessels and a thickness closer to the ground truth. Additionally, in the third row, our model is more successful in generating laser scars compared to other models. Unlike traditional ControlNet, which mistakenly identifies each laser scar as a vessel, our model correctly identifies and transforms the lesions, even with a small dataset. This indicates that our model is particularly advantageous for the transition task with limited data.

\subsection{Ablation Studies}
In this section, we examine the effectiveness of the proposed modules in our approach. We conduct an ablation study by comparing the FID, IS, PSNR, and MS-SSIM metrics with and without the proposed modules and preprocessing strategy. The results of this analysis are presented in Table \ref{tab:ablation_study}.

\textbf{Gated Convolutional Encoder: } To assess the effectiveness of our proposed module, we conducted two experiments. First, we tested the Gated Module, which extracted useful information from the encoder through a gating mechanism. For this evaluation, we removed the Gated Module (denoted as w/o GM), allowing the convolutional information from the encoder to enter the VAE decoder directly without filtering. This removal led to an increase in FID to 82.386, while IS, PSNR, and MS-SSIM decreased from 1.7578, 30.6727, and 0.7104 to 1.6787, 30.0977, and 0.6554, respectively. The second experiment further confirmed the impact of the Gated Convolutional Encoder on generative performance with a small dataset. Without the Gated Convolutional Encoder (w/o GCE), FID increased to 84.5027, and IS, PSNR, and MS-SSIM further declined from 1.7578, 30.6727, and 0.7104 to 1.5496, 28.7481, and 0.6595, respectively. In addition, the reconstruction effects with or without GCE are shown in Fig. \ref{fig:w/o_GCE}. During reconstruction, lacking GCE results in the loss of small blood vessels and details of lesions. The results show that increasing information improves the quality of generated images, but directly incorporating UWF-SLO spatial information without filtering actually decreases the quality. 

\textbf{Cross-temporal Reginal Difference Loss: }
We conducted ablation studies to investigate the impact of CTRD Loss and its hyperparameter $\alpha$ on the model's performance. Initially, we removed the CTRD Loss (denoted as w/o CTRDL), which resulted in an increase in FID from 77.6596 to 80.3336, and a decrease in IS, PSNR, and MS-SSIM from 1.7578, 30.6727, and 0.7104 to 1.6676, 30.1324, and 0.4278, respectively.

Next, we examined the influence of the hyperparameter $\alpha$ by setting it to either 0.25 or 1 throughout the entire training process. When $\alpha$ was consistently set to 0.25, FID slightly increased to 79.9259, with IS, PSNR, and MS-SSIM decreasing from 1.7578, 30.6727, and 0.7104 to 1.6883, 30.3782, and 0.6666, respectively. This indicates that maintaining a constant $\alpha$ value of 0.25 has a minimal impact on generation quality. Conversely, when $\alpha$ was kept at 1, FID rose to 82.5635, and IS, PSNR, and MS-SSIM decreased from 1.7578, 30.6727, and 0.7104 to 1.6260, 28.7534, and 0.5621, respectively. These results suggest that maintaining a high value of $\alpha$ throughout the training process may diminish the model's performance.
\begin{table}[t]
\caption{The upper/lower part represents the ablation in terms of the proposed modules/preprocessing strategies respectively.}
\begin{tabular}{lllll}
\toprule
                & FID($\downarrow$)             & IS($\uparrow$)             & PSNR($\uparrow$)               & MS-SSIM($\uparrow$)          \\  \midrule
w/o GM          & 82.3860           & 1.6787          & 30.0977          & 0.6554          \\
w/o GCE         & 84.5027          & 1.5496          & 28.7481          & 0.6595          \\
w/o CTRDL       & 80.3336          & 1.6676          & 30.1324          & 0.4278          \\
$\alpha = 0.25$         & 79.9259          & 1.6883          & 30.3782          & 0.6666          \\
$\alpha=1   $          & 82.5635          & 1.6260           & 28.7534          & 0.5621          \\
w/o LFEN        & 88.4215          & 1.6170           & 28.6761          & 0.4184          \\ \midrule
w/o ImgS &85.9619 &1.6641  &29.3813  &0.5835 \\
w/o Reg  &104.4215  &1.5167  &25.3922 &0.3601 \\ \midrule
\textbf{LPUWF-LDM(Ours)} &\textbf{77.6596} &\textbf{1.7578} & \textbf{30.6727} & \textbf{0.7104}
\\ \bottomrule
\end{tabular}
\label{tab:ablation_study}
\end{table}

\textbf{Low-Frequency Enhanced Noise: }
We conducted an ablation study to evaluate the importance of Low-Frequency Enhanced Noise (LFEN). When the low-frequency noise was removed (denoted as w/o LFEN), we observed a decline in performance. Specifically, the FID increased to 88.4215, while the IS, PSNR, and MS-SSIM decreased from 1.7578, 30.6727, and 0.7104 to 1.6170, 28.6761, and 0.4184, respectively. The results indicate that the new noise addition strategy could significantly improve the generation quality. The visual difference can be viewed in Fig. \ref{fig:Low Frequency Enhanced noise}. 

\textbf{Preprocessing Strategy: }
In Section \ref{section:preporcessing}, we implement image sharpening for UWF-SLO and perform registration on the early-phase and late-phase UWF-FA. To evaluate the impact of these preprocessing steps, we conducted two experiments, with one using only raw UWF-SLO without image sharpening (designated as w/o ImgS) and another using early-phase and late-phase UWF-FA without registering (designated as w/o Reg). Without image sharpening, the model's performance declined due to some blood vessels blending with the background. The FID increased to 85.9619, and IS, PSNR, and MS-SSIM decreased to 1.6641, 29.3813, and 0.5835, respectively. The sharpening of UWF-SLO makes the difference between blood vessels and background clearer, allowing the model to generate better. 

Second experiment resulted in a marked increase in FID to 104.4215, with IS, PSNR, and MS-SSIM decreasing to 1.5167, 25.3922, and 0.3601, respectively. The results indicate that CTRD Loss has a negative impact on model training on misaligned early and late-phase UWF-FA, resulting in decreased model performance. 
\section{Conclusion}
This paper introduces an advanced latent diffusion framework named LPUWF-LDM, designed to generate high-resolution, detail-rich UWF-FA images from UWF-SLO, specifically targeting the late-phase generation of UWF-FA images. The framework tackles the challenge of producing high-quality UWF-FA images from a small sample dataset by incorporating a CTRD Loss and a low-frequency enhanced noise strategy. These enhancements significantly improve the depiction of blood vessels and pathological details in the generated images. A significant feature of our framework is a Gated Convolutional Encoder is employed in the VAE component. This encoder effectively extracts crucial information from UWF-SLO images and regulates the incorporation of noise information, thereby enhancing the model's expressive power and stability. Experimental results demonstrate that LPUWF-LDM achieves state-of-the-art performance on a proprietary UWF image dataset, offering robust support for non-invasive retinal disease diagnosis.

Looking ahead, we plan to expand the dataset by including a more diverse and complex array of retinal disease cases. Besides, in addition to self-supervised attention to lesions, we will explicitly incorporate some lesion attention mechanisms. Additionally, we aim to integrate clinical information and collaborate closely with ophthalmology experts to develop detailed operating procedures and evaluation standards. These efforts will promote the stable clinical application of our technology and improve the efficiency of retinal disease diagnosis.

\section*{Acknowledgements}
This work was supported in part by the Open Project Program of the State Key Laboratory of CAD\&CG (Grant No. A2410), Zhejiang University, Zhejiang Provincial Natural Science Foundation of China (No. LY21F020017, 2022C03043, 2023C03090 ), National Natural Science Foundation of China (No.61702146, U20A20386, U22A2033),  Chinese Key-Area Research and Development Program of Guangdong Province (2020B0101350001), GuangDong Basic and Applied Basic Research Foundation (No. 2022A1515110570), Innovation teams of youth innovation in science and technology of high education institutions of Shandong province (No. 2021KJ088), the Shenzhen Science and Technology Program (JCYJ20220818103001002), and the Guangdong Provincial Key Laboratory of Big Data Computing, The Chinese University of Hong Kong, Shenzhen.

\printcredits

\bibliographystyle{cas-model2-names}


\end{document}